%% file: main.tex
\documentclass[letterpaper, 10 pt, conference]{ieeeconf}

\IEEEoverridecommandlockouts

\input{macros.tex}

\title{\LARGE \bf ProNav: Proprioceptive Traversability Estimation for Legged Robot Navigation in Outdoor Environments
}

\author{Mohamed Elnoor, Adarsh Jagan Sathyamoorthy, Kasun Weerakoon, Dinesh Manocha \\
\footnotesize{Technical report and video can be found at \url{http://gamma.umd.edu/pronav/}}}

\begin{document}

\maketitle

\begin{abstract}
We propose a novel method, ProNav, which uses proprioceptive signals for traversability estimation in challenging outdoor terrains for autonomous legged robot navigation. Our approach uses sensor data from a legged robot's joint encoders, force, and current sensors to measure the joint positions, forces, and current consumption respectively to accurately assess a terrain’s stability, resistance to the robot’s motion, risk of entrapment, and crash. Based on these factors, we compute the appropriate robot gait to maximize stability, which leads to reduced energy consumption. Our approach can also be used to predict imminent crashes in challenging terrains and execute behaviors to preemptively avoid them. We integrate ProNav with an exteroceptive-based method to navigate real-world environments with dense vegetation, high granularity, negative obstacles, etc. Our method shows an improvement up to 40\% in terms of success rate and up to 15.1\% reduction in terms of energy consumption compared to exteroceptive-based methods.



\end{abstract}


\input{1_Introduction}

\input{2_Related_Works}

\input{3_Background}

\input{4_Instability}

\input{6_Results}
\input{7_Conclusion}

\bibliography{references}
\bibliographystyle{IEEEtran.bst}

\end{document}

%% file: macros.tex
\usepackage{xcolor}
\usepackage{amssymb}
\usepackage{graphicx}
\usepackage{mathtools}
\usepackage{amsmath}
\usepackage{gensymb}
\usepackage{float}
\usepackage{multirow}
\usepackage{makecell}
\usepackage{mathtools}
\usepackage{hyperref}
\hypersetup{
    colorlinks=true,
    linkcolor=blue,
    filecolor=magenta,      
    urlcolor=blue,
    pdftitle={Overleaf Example},
    pdfpagemode=FullScreen,
    }

\usepackage{tabularx}
    \newcolumntype{L}{>{\raggedright\arraybackslash}X}
\usepackage[utf8]{inputenc}
\usepackage[english]{babel}

\usepackage{amsthm}

\usepackage{algorithm}
\usepackage{algpseudocode}

\usepackage{subcaption}

\newcommand{\no}{\noindent}

\newtheorem{probform}{Formulation}[section]


%% file: 1_Introduction.tex
%
%

\section{Introduction}  \label{sec:Intro}
In recent years, autonomous legged robots have found applications in surveillance/monitoring \cite{chen2021autonomous}, exploration \cite{goldberg2002stereo}, and search and rescue \cite{bellicoso2018advances}, etc. The key advantage that enables such applications is their superior capabilities in traversing complex terrains, ones that are inaccessible to wheeled and tracked robots. 

It is important to develop autonomous methods for navigation in complex terrains, which can be broken down into three major categories: uneven/rocky outdoor terrains, dense vegetation, and granular terrains like sand and mud. The uneven or rocky terrains challenge the robot's stability as they often lack solid footholds with sudden variations in elevation \cite{tennakoon2020probe}. Dense vegetation introduces another layer of complexity, presenting risks of entanglement in branches, dried grass, or bushes \cite{sathyamoorthy2023vern,lee2020learning}, leading to unstable behaviors such as slipping and tripping. The third category, granular terrains, often leads to the robot's legs sinking into surfaces like sand or mud due to their deformability under the robot's weight \cite{kolvenbach2019haptic}. Each of these terrain types presents unique difficulties for legged robots, which can affect their navigational capabilities.



\begin{figure}[t]
    \centering
    \includegraphics[width=0.85\columnwidth]{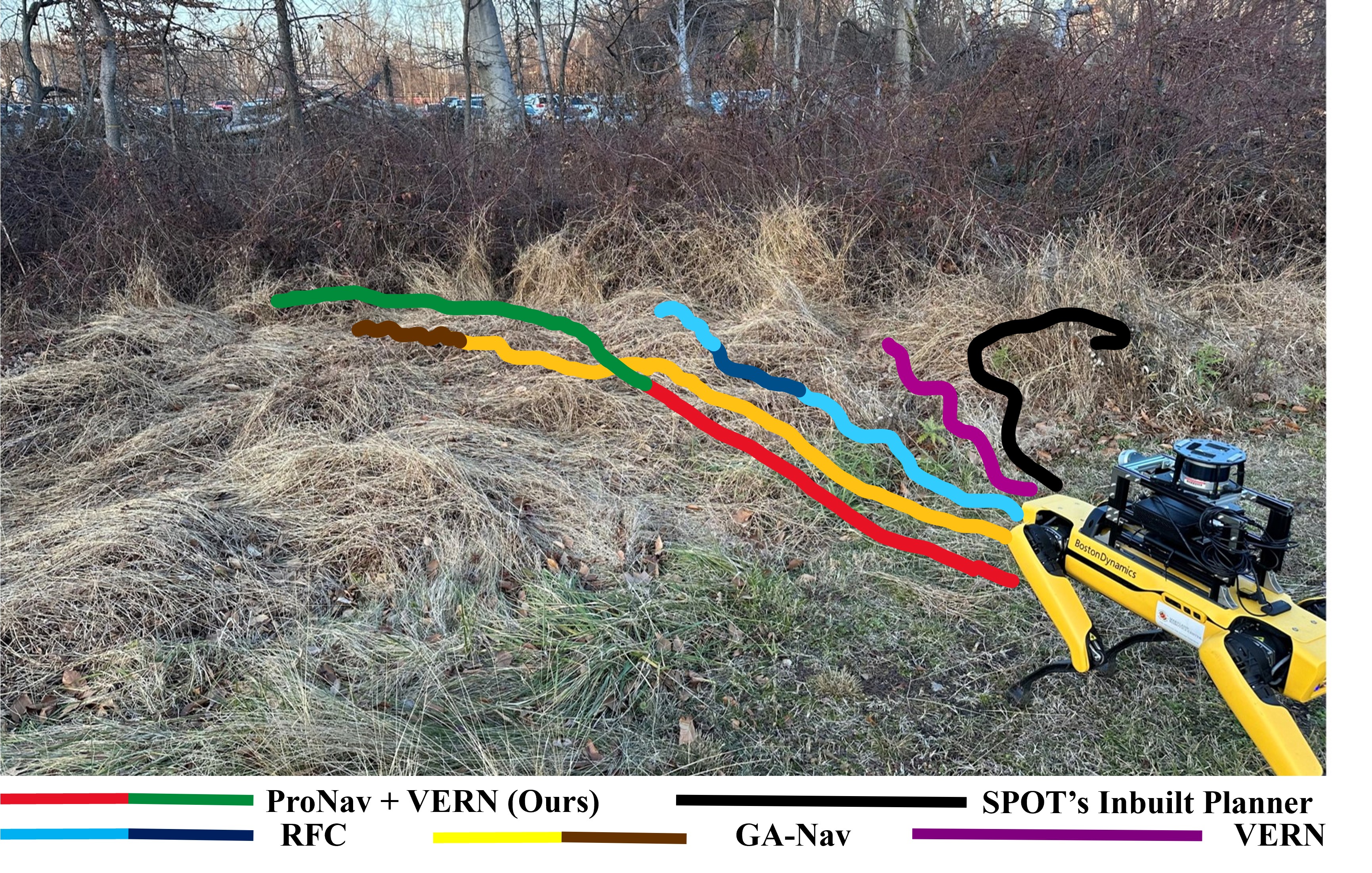}
    \caption{\small{Comparison of our method ProNav with other methods navigating a Spot robot through dense vegetation: ProNav adapts between two gaits: trot (in red), and amble (in green), Spot's in-built planner (black),
    GA-Nav \cite{guan2022ga} (trot: yellow, crawl: brown), RFC\cite{kertesz2016rigidity} (trot: light blue and amble: dark blue), and VERN \cite{sathyamoorthy2023vern} (purple). 
    In this scenario, we observe that our method successfully traverses the dense vegetation due to its efficient gait adaptation and accurate proprioception-based traversability estimation. 
   }}
    \label{fig:cover}
    \vspace{-10pt}
\end{figure}

To tackle these challenges, the robot must be able to accurately evaluate a terrain’s traversability (a measure of the ease of navigation) and then plan its trajectories. Existing methods typically utilize exteroceptive modalities (RGB images, lidar point clouds, and scans) \cite{fahmi2022vital,agarwal2023legged,gennery1999traversability,putz20163d} for traversability estimation. Such exteroceptive methods can provide valuable information about the terrain before walking over it. However, these methods experience degradation in perception accuracy in environments with high occlusions, poor illumination, scarce features, etc. For instance, the terrain geometry could be occluded by dense vegetation. Moreover, certain entities (e.g. negative obstacles such as ditches, and potholes) and changes in a terrain’s properties (dry sand versus wet sand) cannot be accurately detected by exteroceptive modalities.  

To overcome these limitations, several methods have fused exteroception with proprioception to evaluate a terrain’s traversability \cite{fu2022coupling,homberger2019support}. Proprioception measures the state of the robot’s joint and body position and force feedback \cite{al2020review}, while exteroception sensing measures the state of the environment using sensors such as cameras, LiDAR, etc. Although proprioception cannot provide a look-ahead for the terrain, it more accurately represents the robot’s stability on a terrain since unstable walking behaviors are reflected by significant changes in the positions, forces experienced at certain joints, and high energy consumption. Existing research works on proprioceptive traversability analysis have predominantly focused on environments where the robot encounters slippage \cite{carius2019trajectory,teng2021legged,haldane2014detection}, and have not handled regions where the robot's legs could get entangled (e.g. in dense vegetation). 

\begin{figure}[t]
        \centering
        \includegraphics[width=0.8\columnwidth]{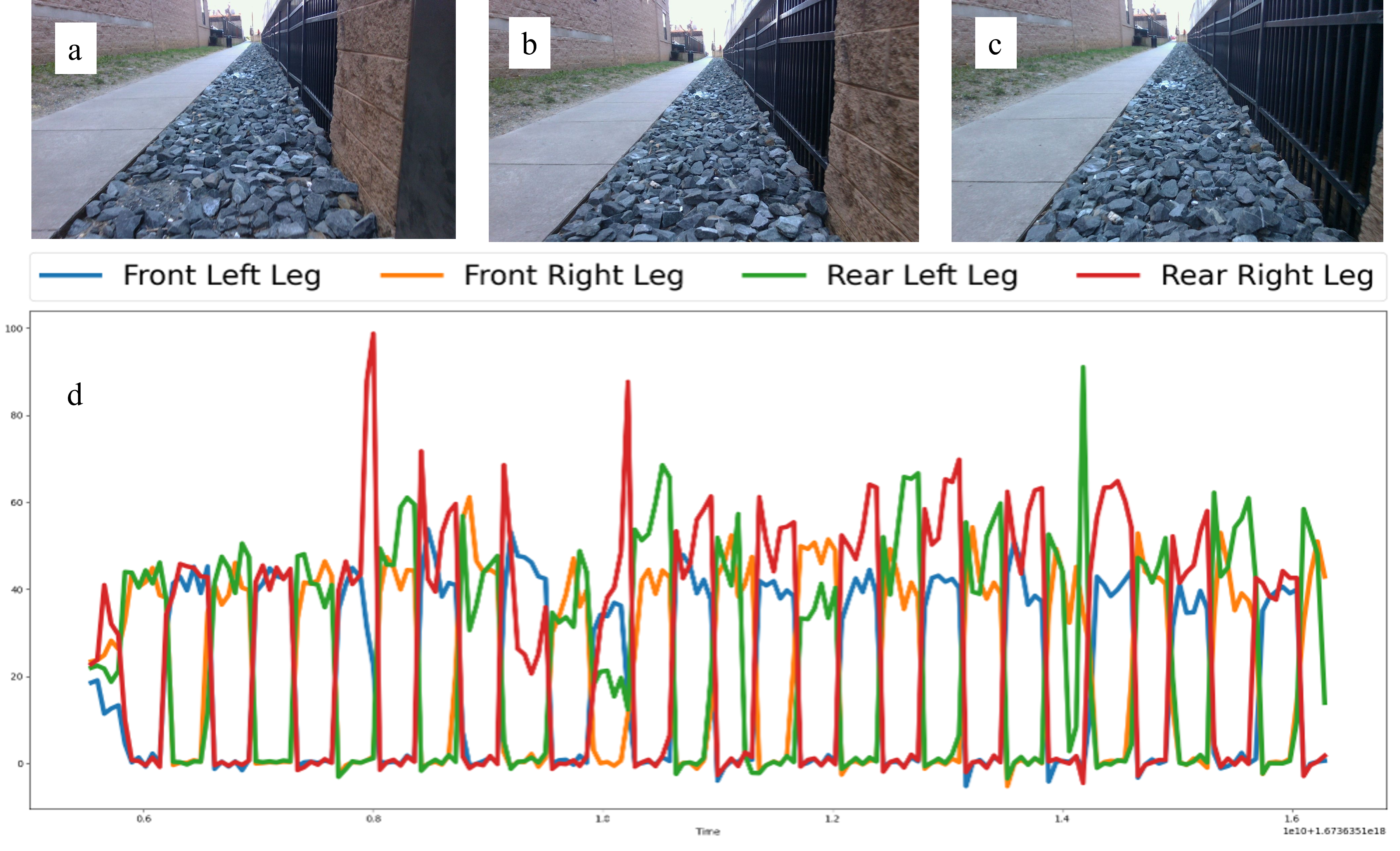}
        \caption{\small{Images (a)-(c) depict the RGB images captured sequentially from the robot's camera. (d) Plot of the fluctuations in knee force readings experienced by the robot while traversing the terrain. The high fluctuations represent instances when the robot became unstable. This shows that visually identical terrains could have different stability characteristics.}}
        \label{fig:rocks}
        \vspace{-10pt}
\end{figure}

Besides that, certain terrains such as concrete and asphalt can be traversed using a single "best” gait. However, this does not apply to all terrains. For example, a grassy terrain may appear uniform but can vary significantly, transitioning from dry to muddy areas with similar visual appearances. Similarly, navigating rocky terrain presents a similar set of challenges as shown in Figure \ref{fig:rocks}. These situations indicate that a legged robot must adapt its gait based on proprioceptive feedback instead of only following visual sensing.


\textbf{Main Contributions:} To address these limitations, we propose ProNav, an approach for using proprioception for improved terrain traversability estimation in a variety of environments (rocky, granular, densely vegetated, etc). The proprioceptive signals are measured from a legged robot’s joint encoders, force, and current sensors.  The novel components of our work include:

\begin{itemize}
  \item A novel terrain traversability estimation method using only proprioceptive signals (joint positions, forces, current consumption) to characterize the stability, and resistance to the robot's motion on a terrain. Our method uses the aforementioned signals to estimate traversability using Principal Component Analysis (PCA) within $1$ second of walking on a new terrain type using edge computing hardware with limited computation power.

  \item A novel crash prediction mechanism that can foresee slipping, tripping, and leg entrapment-related crashes. 
  This leads to an improvement of 40\%  in terms of success rate in densely vegetated regions where all other methods experienced difficulties in reaching the goal.

  \item A novel gait adaptation approach that selects the appropriate gait leading to increased stability (lower vibrations), and lower energy consumption while traversing challenging terrains. We highlight ProNav's performance by integrating it with an exteroception-based navigation method for traversing through dense vegetation, and rocky and granular terrains.
  
\end{itemize}

%% file: 2_Related_Works.tex
\section{Related Works}

In this section, we discuss the existing methods for estimating terrain traversability. Next, we analyze the existing navigation and planning techniques for legged robots.

\subsection{Perception for Navigation}
Autonomous robot navigation in challenging environments requires robots to perceive the real world through their sensors. To this end, robots often incorporate onboard exteroceptive, and proprioceptive sensors. We briefly review the existing work on exteroceptive and proprioceptive perception in the following sub-sections.


\subsubsection{Exteroceptive Sensors}
A popular approach is the use of geometry-based methods which reconstruct a 3D representation of the environment by using technologies such as LiDAR or stereo cameras \cite{gennery1999traversability} . 
Another approach, as presented in \cite{putz20163d}, generates a 3D triangle mesh of the environment from a 3D point cloud, which is then input into an online path planner for local navigation. 
Recently, \cite{frey2022locomotion} proposed learning terrain traversability by training a sparse 3D network of occupancy maps.
However, these geometry-based methods have limitations, including difficulties with deformable surfaces such as sand, obstacles like tall grass, and the risk of poor estimation \cite{gennery1999traversability,goldberg2002stereo}.

Concurrently, vision-based approaches have seen widespread application in robot perception \cite{fahmi2022vital,agarwal2023legged}. Previous work in semantic segmentation categorizes terrain properties into traversable and non-traversable classes. For instance, Guan et al \cite{guan2022ga} leverage a multi-head vision transformer architecture to segregate the terrain into six distinct categories. 
Also, traversability classification can be performed using anomaly detection from multi-model images \cite{wellhausen2020safe}. Even though such vision-based systems perform well under perfect weather conditions, they often result in erroneous classification due to 
lighting changes \cite{aladem2019evaluation}.

Several studies have also explored the potential of sensor fusion for terrain classification \cite{schilling2017geometric, wisth2022vilens,weerakoon2022graspe}. Notably, in \cite{schilling2017geometric}, geometric and vision-based techniques are used to deliver improved performance.  In \cite{weerakoon2022graspe}, reliability-aware sensor fusion is performed to mitigate the performance degradation due to cluttered sensing. Recently, \cite{sathyamoorthy2023vern} proposed VERN, which utilizes a lightweight Siamese network to classify complex outdoor vegetation based on traversability.
The method in \cite{sathyamoorthy2022terrapn} employs IMU sensor data to learn surface traction, bumpiness, and deformability using an online self-supervised learning strategy. 
While this approach has shown promising results for a number of terrains, others like rocks and bushes, with irregular texture/structure, were not investigated.

\subsubsection{Proprioceptive sensors}
In outdoor environments, exteroceptive sensors could receive noisy data because of factors such as degraded lighting conditions and occlusions. Also, the environment can be extreme and challenging. For instance, the ground could be covered by vegetation (e.g., short/tall grass, bushes) and the robot cannot recognize the terrain type using vision or LiDAR. To overcome such issues, there has been a continuous development in proprioceptive perception \cite{miki2022learning}. Moreover, proprioception can be coupled with vision in legged robots as in \cite{fu2022coupling}, where Fu et al. use the camera to create a cost map around the robot, while the terrain traversability is mainly evaluated based on proprioceptive feedback. That also helps in avoiding unexpected obstacles such as glass walls. \cite{loquercio2023learning} proposed a cross-modal algorithm that uses an RGB camera and shifted proprioception to learn a walking locomotion policy.
More recently, Dey et al. \cite{dey2022prepare} leverage the proprioceptive information from a legged robot's joints to predict slip and fall events with high accuracy. However, the robot is operated in a limited number of terrains such as rubble and other uneven, underground terrains, and not in densely vegetated environments. Moreover, their proposed model primarily predicted slipping and tripping and it is not used for navigation. Our novel approach uses proprioceptive feedback and current consumption from the actuators to also detect entanglement in dense vegetation and recover the robot.

\subsection{Outdoor Navigation} 

Recently, many approaches have been proposed to leverage the agile mobility of legged robots \cite{truong2023indoorsim,wellhausen2023artplanner} in unstructured outdoor environments, which is challenging for wheeled robots \cite{biswal2021development}.
Some of these works use cost maps to represent the traversability of the environments \cite{overbye2021path}.Semantic Belief Graph are utilized in \cite{ginting2023safe} to train a policy for trajectory generation in extreme environments. Moreover, a traversability uncertainty-based method is proposed in \cite{guzzi2019impact}. In \cite{garcia2022gait}, the authors presented a traversability estimator that uses a classifier (or a regressor) neural network based on elevation maps.
Artplanner \cite{wellhausen2023artplanner} is a navigation planner designed for the DARPA Subterranean Challenge that uses geometric reachability checking and a motion cost neural network to compute optimal paths.
Proprioceptive feedback is also used in the literature \cite{lee2020learning,fu2022coupling,dey2022prepare}. 
In \cite{lee2020learning}, Lee et al. utilized proprioceptive feedback to train a robot controller using reinforcement learning. Their approach shows zero-shot capabilities when tested in outdoor settings. 
However, an inherent limitation of proprioception is its inability to preview terrain features before the robot directly interacts with them. This limitation motivates the integration of ProNav with an exteroceptive-based navigation method, ensuring a more comprehensive navigation strategy.

%% file: 3_Background.tex

\section{Background}
In this section, we explain our assumptions, define important notations used, and our problem formulation. 

\subsection{Setup and Conventions}
We assume a quadrupedal robot with 12 degrees of freedom (DOF), with 2-DOFs in the hip, and 1-DOF in the knee of each leg. We assign numbers 1, 2, 3, and 4 to denote the front-left, front-right, rear-left, and rear-right legs respectively, and $i$ to denote each leg. A robot coordinate frame is established at its center of mass with positive X, Y, Z pointing forward, left, and up respectively. Frames with similar conventions are established at each hip and knee joint. 
The hip has two actuators, one is moving along the X-axis direction and the other one along the Y-axis direction. Moreover, the knee actuator moves along the Z direction. We also measure the positions $p^{hip,i}_{x}, p^{hip,i}_{y}$, $p^{knee,i}_{z}$, velocity $v^{hip,i}_{x}, v^{hip,i}_{y}$, $v^{knee,i}_{z}$, and force $f^{hip,i}_{x}, f^{hip,i}_{y}, f^{knee,i}_{z}$ exerted at a time instant $t$. Several widely used legged robot platforms possess these specifications and capability to measure these parameters \cite{anymal,ghost-robot,BostonDynamics2024}. 

Position and velocity data at the joints are measured using encoder sensors, and the forces experienced are measured using the internal tactile sensing mechanism. Finally, we assume that the current drawn ($I(t)$) from the robot's battery can be measured using an ammeter or a current sensor while traversing various terrains. We define $\mathbf{X}_t \in \mathbb{R}^{36}$ as the set of all positions (3), velocities (3), and forces (3) obtained from all four legs of the robot. Based on our setup and notation, we have formulated the state vector at a given time instant $t$ for our traversability estimation method as,

\begin{equation}
    \text{State Vector} = \left[ \mathbf{X}_t \in \mathbb{R}^{36}, I(t) \in \mathbb{R} \right].
\end{equation}


\subsection{Problem Domain}
The focus of our approach is to enhance the navigational capabilities of legged robots traversing through a variety of terrains (e.g. densely vegetated, granular, rocky) using proprioceptive feedback to 
adapt to changes in surface conditions. In these terrains, the robot's legs could slip, trip, sink, or get entangled. A robot falling to the ground (we define as a \textit{crash}) which could be caused by one of the following reasons: 

\no \textbf{Poor Foothold}: This causes the robot's feet to slip in rocky or slippery terrains because the robot's feet do not have a firm, flat surface to support themselves on. 
    
\no \textbf{Granularity}: This causes the robot's feet to sink into the terrain (e.g. sand, mud, snow) leading to erroneous measurements of joint positions. This could cause the robot's controller to overcompensate to stabilize itself. 
    
\no \textbf{Resistance to Motion}: This is typically caused by dense, pliable vegetation that can be passed through (e.g. tall grass and bushes). Additionally, the robot's legs could get entangled with vegetation causing higher resistance to motion.

\begin{figure*}[t]
    \centering
    \includegraphics[width=0.85\linewidth]{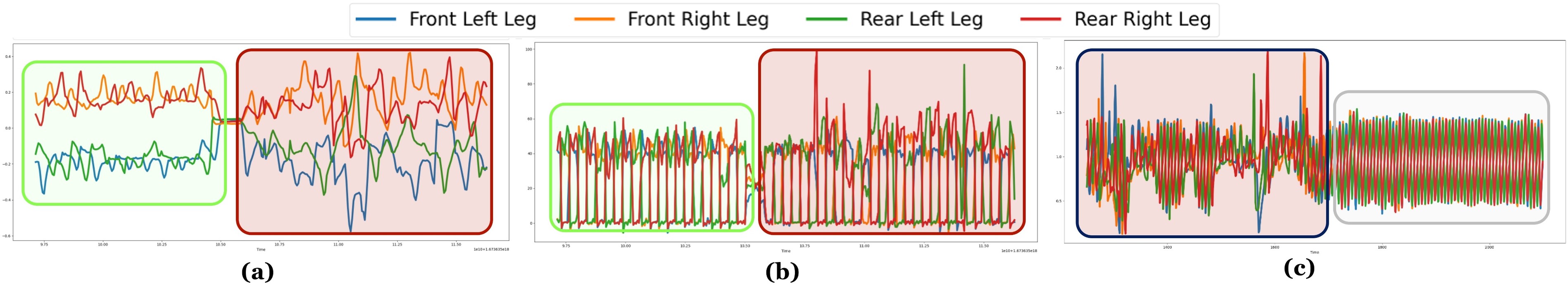}
    \caption{\small{\textbf{(a)} Changes in the hip X-axis position of the robot while traversing grass (green box), and rocks (brown box) plotted over time. \textbf{(b)} Force exerted by the four knee actuators while traversing grass (green box), and rocks (brown box). Steady readings observed on stable grass terrain reflect ease of traversal, while the increased volatility and noticeable spikes on the rocky terrain are indicative of increased resistance and slippage, causing variable load on the actuators. \textbf{(c)} Changes in the hip Y-axis position of the robot while traversing dense vegetation (violet box), and concrete (gray box). High fluctuations are observed while traversing dense vegetation due to the legs' entanglement instances. Conversely, a steady and consistent reading is observed during concrete traversal.
   }}
\label{fig:prop-signal-analysis}
\end{figure*}






To traverse various terrains, we assume a legged robot with a locomotion model that can alternate between three gaits: trot, crawl, and amble \cite{dey2022prepare,truong2023indoorsim}. Trot is the standard walking gait where the robot walks with two of its feet on the ground at a time instance, allowing fast movements. It is stable on hard surfaces, with moderate power consumption. On the other hand, during crawl and amble, the robot has three of its feet on the ground at a time instance, leading to more stable behaviors in uneven, granular, deformable surfaces. Amble helps to traverse through environments with high resistance to motion while also maintaining stability, which also helps handle poor foothold terrains. Similarly, crawl maintains high stability in granular terrains and regions with poor footholds while consuming minimal power. The maximum velocities for each gait follows the trend $v^{trot}_{max} = v^{amble}_{max} > v^{crawl}_{max}$, and the current consumption for each gait follows $I^{Amble} > I^{Trot} > I^{Crawl}$. Based on these definitions, our formulation can be stated as follows,

\begin{probform}
    To adaptively select a stable gait $g^*$ given collision-free, goal-directed velocities $(v^*, \omega^*)$, by assessing a terrain's traversability based on a set of proprioceptive signals from a legged robot to improve stability and prevent crashes.
\end{probform}

%% file: 4_Instability.tex
\section{ProNav: Proprioception-based Stable Navigation}
In this section, we analyze and choose the relevant proprioceptive signals, process them to assess stability, and explain our gait adaptation strategy to stabilize the robot.

\subsection{Analysis of Proprioceptive Signals}
Our goal is to choose the fewest number of proprioceptive signals (i.e., the minimum subset $\mathbf{Y}_t \subset \mathbf{X}_t$ at every time instant $t$) that are also excellent indicators of stability. Deducing the minimum subset helps reduce the input dimensionality of our approach, which in turn improves its real-time factor.

\no \textbf{Hip's Position:} Our empirical analysis revealed a strong correlation between the amount of slip on a terrain and the change in hip position along the X-axis and Y-axis. Figure \ref{fig:prop-signal-analysis} visually represents these changes as the robot navigates three different types of terrain, each representing different levels of traversability: dense vegetation, rocks, concrete, etc.


\no \textbf{Knee's force:} Sudden peaks in the forces experienced by the robot's knee actuators (Fig. \ref{fig:prop-signal-analysis}) along the Z direction indicate an absence of stable footholds due to unevenness, causing the robot to exert more effort to stabilize itself. 



\begin{figure}[h]
    \centering
    \includegraphics[width=\columnwidth,height=3.8cm]{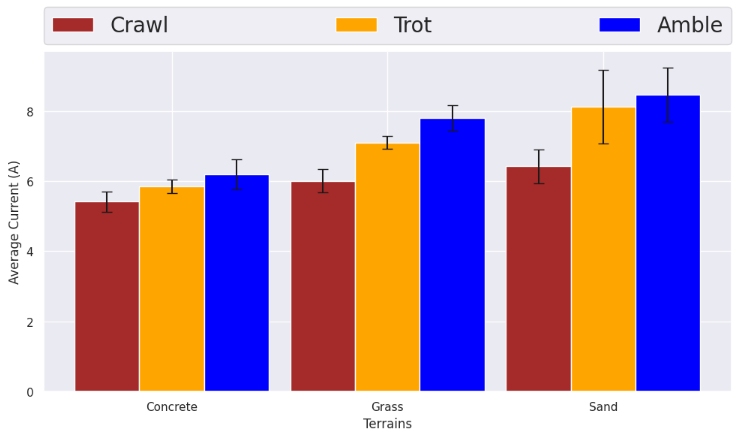}
    \caption{\small{The average current consumption in amperes, with 95\% confidence interval as the robot traverses concrete, grass, and sand. Lower current consumption on concrete indicates ease of traversal. However, higher values on sand highlight increased resistance and energy usage. Additionally, a consistent trend in current consumption is exhibited while using crawl, trot, and amble on various terrains. }}
    \label{fig:current_comp}
    \vspace{-10pt}
\end{figure}

\no \textbf{Current Consumption:} The amount of current consumed while traversing various terrains at a consistent elevation is proportional to the resistance to motion experienced in each terrain 
(Fig. \ref{fig:current_comp}). 
Also, the robot's gait consistently impacts current consumption on different terrains as mentioned before.



\subsection{Prepossessing Proprioceptive Signals}



Our goal in preprocessing the chosen force and position data is to obtain quantities that change drastically on various terrains, thus indicating their properties. Vectors of the processed data are then analyzed using Principal Component Analysis (PCA). 

\subsubsection{Preprocessing Force Data}
At any time instant $t$, we consider the past $n$ samples of $f^{knee,i}_z$ of the $i^{th}$ leg. That is, we consider the vector $[f^{knee,i}_z(t), f^{knee,i}_z(t-1), ... , f^{knee,i}_z(t-n+1)]$. Next, we obtain the mean force for the $i^{th}$ leg's knee as $\mu^i_f = (\sum_{j=0}^{n-1} f^{knee,i}_z(t - j))/n$, and then the mean force experienced by the robot as a whole as $\mu^{rob}_f = (\sum_{i=1}^{4} \mu^i_f)/4$. Finally, we calculate the difference $\Delta^i_f = \mu^{rob}_f - f^{knee,i}_z(t)$ for each leg. As the robot walks on various terrains, $\Delta^i_f < 0$ indicates that the robot has entered a poorly traversable terrain which leads to high knee forces, and $\Delta^i_f > 0$ indicates a highly traversable terrain. To further amplify changes in traversability, we use $\sum_{i=1}^{4} \Delta^i_f$, and a counter that denotes the number of spikes in the force experienced, defined as: $count = count + 1$ if $\Delta^i_t < 0, i \in \{1, 2, 3, 4\} $. 


\subsubsection{Preprocessing Position Data}

At time instant $t$, we consider the past $m$ samples of $p^{hip,i}_x$. For each leg $i$, we calculate the maximum $max_i$ and minimum $min_i$ values of these $m$ samples and finally calculate $\Omega^{rob}_{p,x}(t) = \sum_{i=1}^4 \left| max_{i,x} - min_{i,x} \right|$. $\Omega^{rob}_{p,x}$ represents the magnitude of variation in the hip positions along the X direction. Similarly, we obtain $\Omega^{rob}_{p,y}$ along the Y direction. A high value of $\Omega^{rob}_{p,x}(t)$ or $\Omega^{rob}_{p,y}(t)$ indicates the unavailability of stable footholds which leads to slippage (e.g. in rocky terrains), or the presence of a granular surface that leads to sinkage. 

\subsubsection{Processed Input Vector}
We combine the processed quantities in knee forces and hip positions with the current drawn from the robot's battery to construct the input vector $A \in \mathbb{R}^9$ to estimate terrain traversability as,

\begin{equation}
    A(t) = [\Delta^1_f, \, \Delta^2_f, \, \Delta^3_f, \, \Delta^4_f, \, \sum_{i=1}^{4} \Delta^i_f, \, count, \, \Omega^{rob}_{p,x}, \Omega^{rob}_{p,y} , \, I].
    \label{eqn:input-vector}
\end{equation}

\no All the quantities on the right in equation \ref{eqn:input-vector} are functions of $t$. It is omitted for readability.

\subsection{Terrain Traversability Estimation}
To estimate a terrain's traversability using our preprocessed proprioceptive signals, we first apply Principal Component Analysis (PCA) to reduce its 9 dimensions into two principal components as,

\begin{equation}
    \mathbf{p}_{t} = \text{PCA}(A(t)).
\end{equation}

\begin{figure}[t]
\centering
\includegraphics[width=7cm]{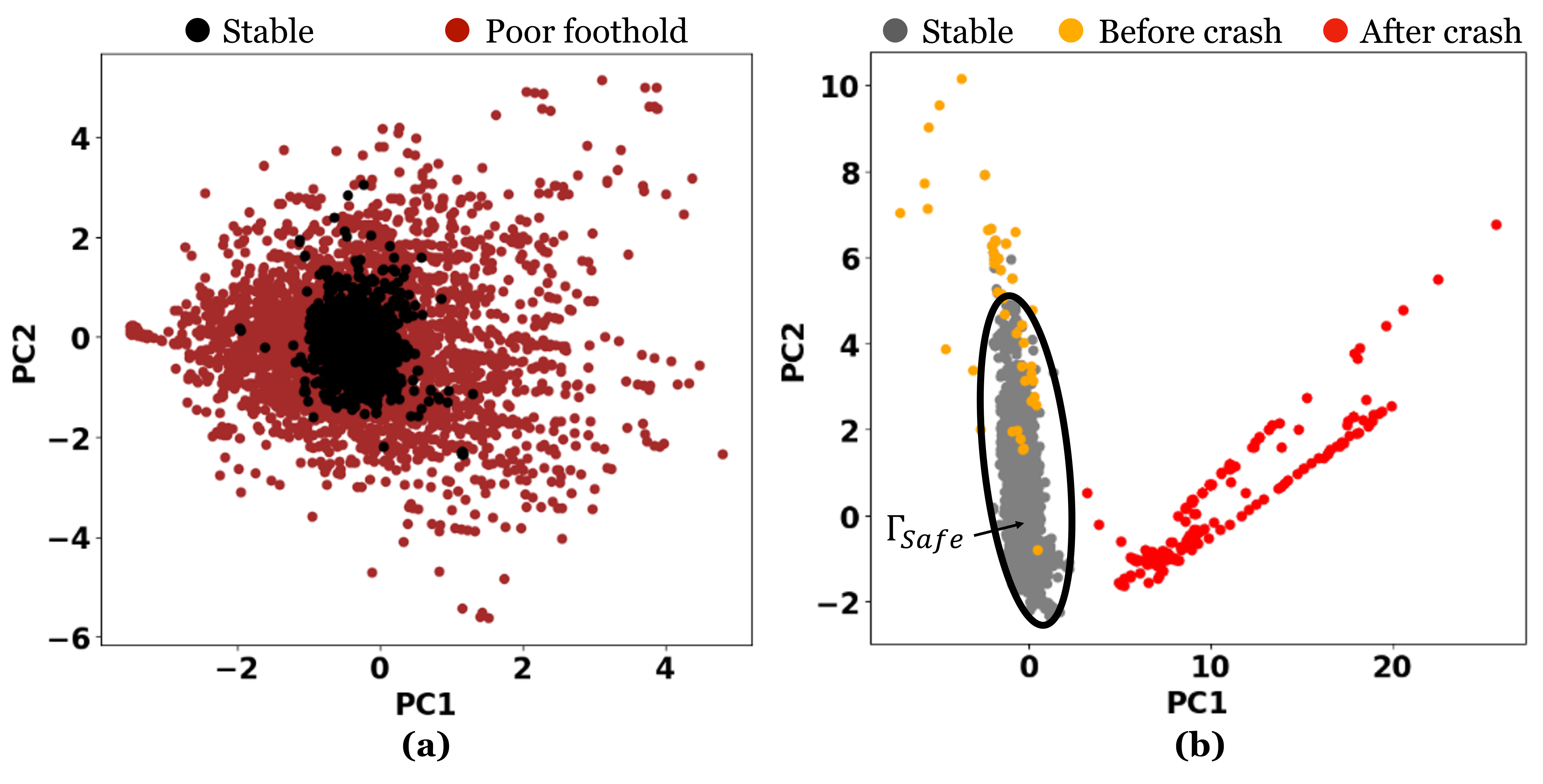}
\caption{\small{(a) PCA applied to key proprioceptive metrics (hip actuator positions, knee actuator force, and battery current) across two different terrains when using the trot gait. The variances along the two principal components indicate the level of stability on a terrain. 
(b) The figure shows the shift in the PCA distribution between stable navigation (grey points), before a crash (yellow), which represents 3 seconds before the crash, and 10 seconds after a crash (red), where a robot falls to the ground. If the robot's proprioceptive signals lie outside the ellipse $\Gamma_{safe}$, the robot is heading towards a crash.}}
\label{fig:pca}
\end{figure}

\begin{figure}[h]
        \centering
        \includegraphics[width=\columnwidth,height=4.0cm]{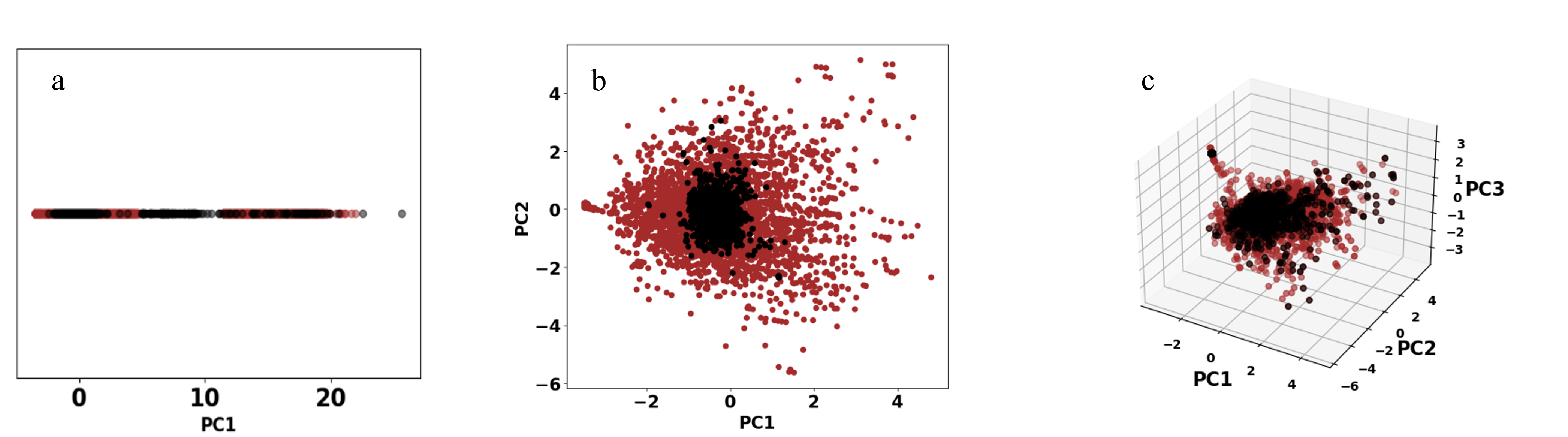}
        \caption{\small{Figure (a) illustrates the use of one principal component, (b) shows the use of two components, and (c) displays the use of three components. In these figures, red indicates the proprioceptive data recorded on rocky terrain, while black denotes concrete terrain while the robot used the trot gait. Notably, the use of two components, as depicted in (b), offers a more distinct and clearer representation of the terrains compared to using either one or three components.}}
        \label{fig:PCA_number}
        \vspace{-10pt}
    \end{figure}

\no Here, $\mathbf{p}_t$ is a 2D point in the PCA space (Fig.\ref{fig:pca}). PCA allows us to simplify and effectively compare different terrains based on these components. We chose to use two principal components because it yielded all the required information needed for traversability estimation. Using just one component was insufficient, and three components did not add new useful information 
in terms of visualizing distributions for each type of terrain as shown in Figure \ref{fig:PCA_number}.

Continuously plotting the PCA points corresponding to traversing a terrain $T$ with gait $g$ for a time period results in a distribution/cluster of points as shown in Fig. \ref{fig:pca}a. We obtain several key insights from our analysis: 1. The variance of the data along the two principal components differentiates stable (low variance/small cluster) and unstable (high variance/big cluster) terrains, 2. Terrain-gait pairs that have similar stability characteristics have similar clusters (e.g. concrete-trot and asphalt-trot), and 3. The position of the PCA points can also aid in predicting imminent crashes with noticeable shifts during the moments immediately before and after crashes (see Fig. \ref{fig:pca}b).

\begin{figure}[htbp]
\centering
\includegraphics[width=\columnwidth]{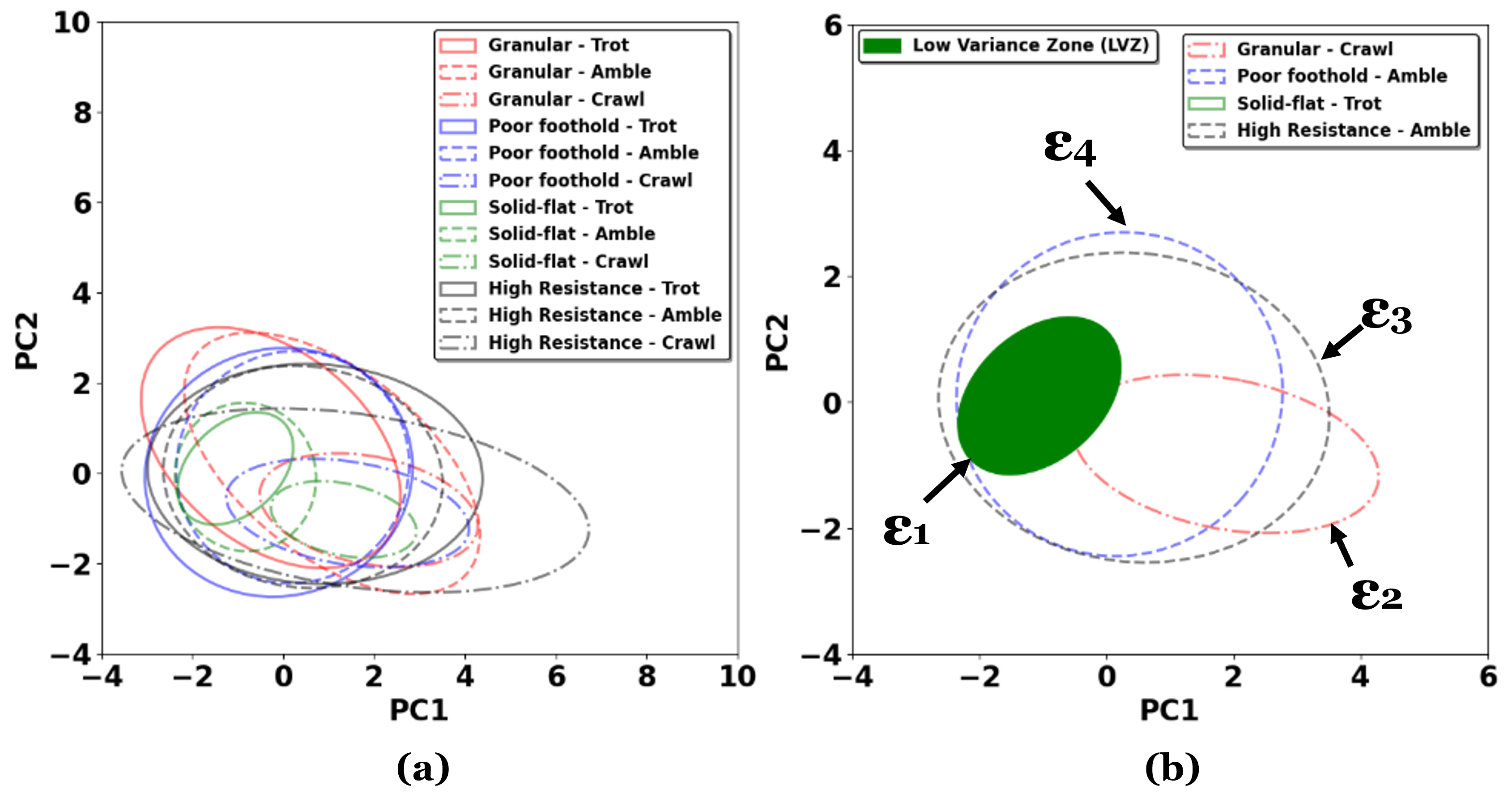}
\caption{\small{The clusters' ellipses for the PCA components of four different terrains, granular is red, poor foothold is blue, solid-flat is green, and high resistance is black. For each terrain, three types of gait data are shown using different ellipse boundaries. The solid line denotes trot, the dashed line denotes amble, and the dash-dot line is for crawl. (a) All 12 ellipses 
(b) The stable zone's (SZ) ellipses $\mathcal{E}_1 - \mathcal{E}_4$, $\text{where SZ} = \left( \mathcal{E}_1 \cup \mathcal{E}_2 \cup \mathcal{E}_3 \cup \mathcal{E}_4 \right) \subset \Gamma_{\text{safe}}$.}}
\label{fig:ellipses}
\end{figure}

We extend this analysis to using all three gaits on terrains with poor footholds, granularity, and high resistance to motion, and obtained a unique cluster of points for each terrain-gait pair. By fitting a 2D Gaussian to each cluster, we obtain a characteristic ellipse (see Fig. \ref{fig:ellipses}a) that forms the boundary of the cluster. Similar to our previous insights, the size/area of each ellipse denotes the robot's stability on a terrain while using a certain gait.


Since our objective is to maintain high stability in all terrain types, while also maintaining a fast progress towards the robot's goal while navigating, we consider only the ellipses with the lowest area to maximum velocity of the gait ratio (Fig. \ref{fig:ellipses}b). We refer to them as high stability ellipses for each terrain type. Of these ellipses, we observe that trotting on stable, flat terrains such as concrete/asphalt creates the ellipse with the smallest area, and highest stability. We refer to this ellipse and its enclosing region as the Low Variance Zone (LVZ), highlighted in Fig. \ref{fig:ellipses}b. Ideally, the current PCA point $\mathbf{p}_t$ indicating the robot's stability should lie within the LVZ. However, on other challenging terrains, $\mathbf{p}_t$ would most likely lie outside the LVZ. Next, we detail how $\mathbf{p}_t$ and the other ellipses in Fig. \ref{fig:ellipses}b can be used to select a stabilizing gait when $\mathbf{p}_t \notin LVZ$ without any exteroceptive feedback.



\subsection{Stable Gait Adaptation}
A key insight from Fig. \ref{fig:ellipses} is that when the appropriate gait for a terrain is chosen (e.g. crawl for granular terrains) at time $t$, the point $\mathbf{p}_{t+i} \, \forall i > 0$ would be contained within the corresponding high stability ellipse (e.g. granular-crawl ellipse) as represented in Fig. \ref{fig:ellipses}b. Conversely, if an inappropriate gait is chosen, $\mathbf{p}_{t+i} \forall i > 0$ will not lie within the high stability ellipse for that terrain. A key challenge is selecting the appropriate gait \textit{without knowing the terrain type}.

To determine the most appropriate stabilizing gait $g^*$ (with the corresponding ellipse $\mathcal{E}^*$) for a terrain, we only consider the four ellipses and their enclosing areas in Fig. \ref{fig:ellipses}b, and refer to them as $\mathcal{E}_1 = LVZ, \mathcal{E}_2, \mathcal{E}_3, \mathcal{E}_4$, as marked. We refer to their union as the Stable Zone ($SZ$). Let us consider two points in subsequent time instances $\mathbf{p}_t$ and $\mathbf{p}_{t+1}$. At time instant $t$, if $\mathbf{p}_t \notin LVZ$, and $\mathbf{p}_t$ lies in the intersection of any of the other ellipses, we calculate the minimum area ellipse as $\mathcal{E}^{MA}_t = argmin_i(area(\mathcal{E}_{i})), \forall \mathcal{E}_{i} \in \{\text{Ellipses in the intersection}\}$. We set $\mathcal{E}^*_t = \mathcal{E}^{MA}_t$, and $g^*_t$ as the associated gait calculated as $gait(\mathcal{E}^*_t)$. If $g^*_t$ is the appropriate gait for the terrain, $\mathbf{p}_{t+1} \in \mathcal{E}^*_t$, and the robot can continue to execute the same gait, i.e., $g^*_{t+1} = g^*_t$.

However, if $g^*_t$ is not the appropriate gait, then $\mathbf{p}_{t+1} \notin \mathcal{E}^*_t$. This leads to two scenarios: $\mathbf{p}_{t+1} \in SZ$, or $\mathbf{p}_{t+1} \notin SZ$. If $\mathbf{p}_{t+1} \in SZ$, then we can compute the minimum area ellipse as before, $\mathcal{E}^{MA}_t = argmin_i(area(\mathcal{E}_{i})), \forall \mathcal{E}_{i} \in \{\text{Ellipses in the intersection}\}$. If $\mathbf{p}_{t+1} \notin SZ$, we select a high stability ellipse based on its distance from $\mathbf{p}_{t+1}$. That is, $\mathcal{E}^{MD}_{t+1} = argmin_i(dist(\{\mathcal{E}_{2}, \mathcal{E}_{3}, \mathcal{E}_{4}\} - \{\mathcal{E}^*_t\}, \mathbf{p}_{t+1}))$. In both cases, $g^*_{t+1} = gait(\mathcal{E}^*_{t+1})$. We temporarily remove $\mathcal{E}^*_t$ from consideration, as the gait corresponding to it caused $\mathbf{p}_t$ to leave the stable zone $SZ$ at time $t$. Intuitively, removing the ellipse allows the formulation to converge to the correct ellipse corresponding to the current terrain type, and its associated gait. The ellipse is added back to consideration for future gait calculations when a gait change is required. To summarize, 

{\small
\begin{equation}
    g^*_{t+1} = 
    \begin{cases}
        \text{Trot} & \text{if } \mathbf{p}_{t+1} \, \in \, \text{LVZ} \text{, }\\
        
        g^*_t &  \text{if } \mathbf{p}_{t+1} \, \notin \, \text{LVZ, } \mathbf{p}_{t+1} \, \in \, \mathcal{E}^*_t \text{, }\\
        
        gait(\mathcal{E}^*_{t+1}) &  \text{if } \mathbf{p}_{t+1} \, \in \, \text{SZ, } \mathbf{p}_{t+1} \, \notin \, \mathcal{E}^*_t \text{, } \mathcal{E}^*_{t+1} = \mathcal{E}^{MA}_{t+1}\\

        gait(\mathcal{E}^*_{t+1}) &  \text{if } \mathbf{p}_{t+1} \, \notin \, \text{SZ, } \mathcal{E}^*_{t+1} = \mathcal{E}^{MD}_{t+1}\\
        
        \text{None} & \text{if } \mathbf{p}_{t+1} \, \notin \, \Gamma_{\text{safe}}.
    \end{cases}
    \label{eqn:gait-optimal}
\end{equation}
}

We also noted a significant pattern in the behavior of $(pc_1(t), pc_2(t))$ leading up to a crash. Specifically, for three seconds before a crash, there is a noticeable shift along the PC2 axis (Fig. \ref{fig:pca}b). Also, the data distribution preceding a crash exhibits high variances, indicating a lack of stability. Based on this observation, we adopted a preventative measure of halting the robot in these scenarios to mitigate the risk of crashes. 

\begin{figure}[t]
    \centering
    \includegraphics[width=0.75\columnwidth,height=3.0cm]{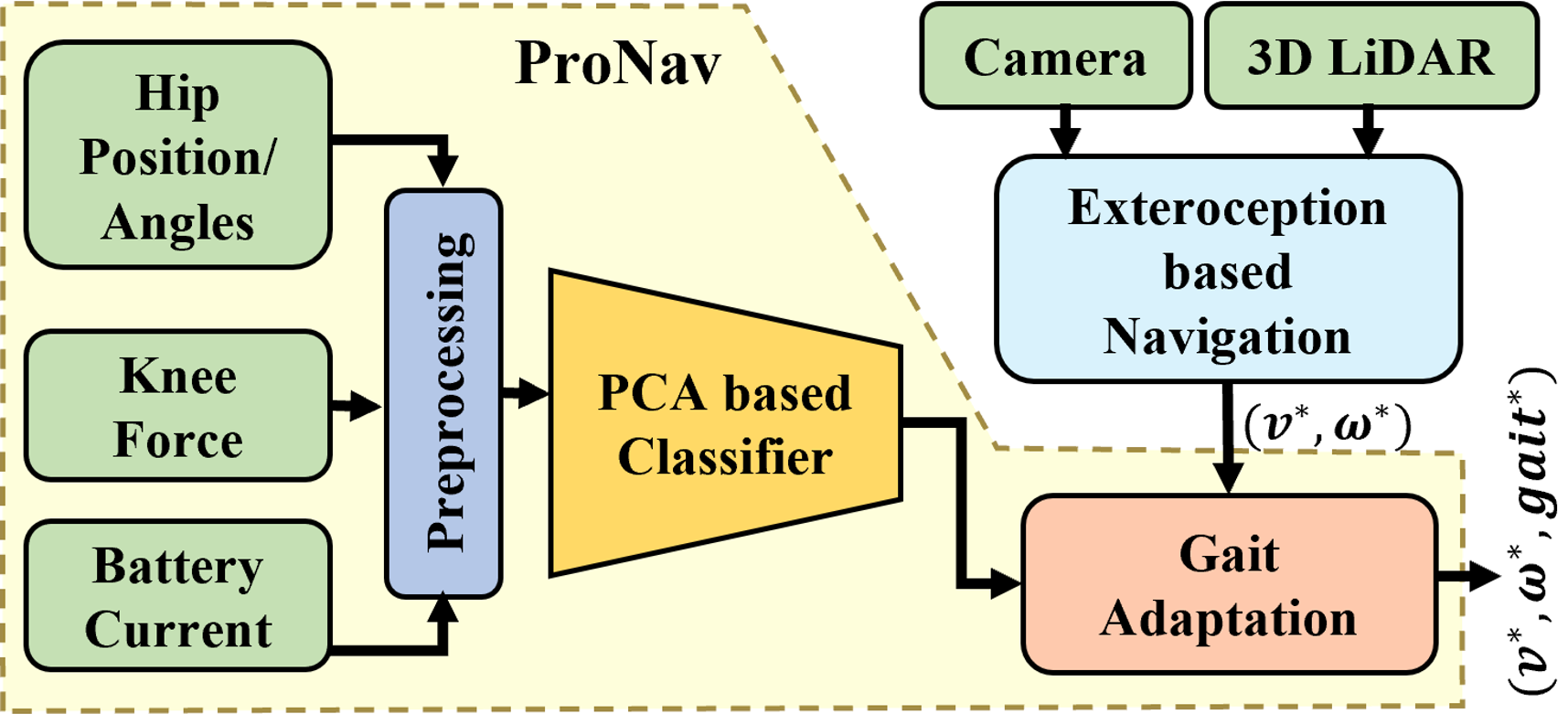}
    \caption{\small{The overall system architecture integrating ProNav with an exteroception-based planner. We utilize hip X- and Y-axis positions, knee force, and current drawn as our proprioceptive signals. Our PCA-based approach estimates the traversability of the terrain, and the gait adaptation selects the improved gait for stability. The camera and lidar are used with the integrated exteroception-based planner for obstacle avoidance.}}
    \label{fig:sys_arch}
    \vspace{-10pt}
\end{figure}

\subsection{ProNav + Exteroception}
Although proprioceptive modalities help to estimate the traversability of the terrain the robot is walking on, they lack \textit{look-ahead} capability (whenever terrain is visible) that exteroceptive sensing affords. That is because the traversability of terrains ahead of the robot cannot be assessed using past and current proprioceptive data. Therefore, fusing exteroception and proprioception helps bring out the best of both worlds. 

ProNav can be easily combined with any navigation method that uses exteroceptive sensing such as RGB images, 3D point clouds, etc. An overview of such a hybrid architecture is shown in Fig. \ref{fig:sys_arch}. The collision-free, goal-directed velocities $(v^*, \omega^*)$ are extracted from an exteroception-based planner \cite{fox1997dynamic,weerakoon2022graspe,sathyamoorthy2023vern}, and the gait evaluated to be the most stable for the current scenario by ProNav (equation \ref{eqn:gait-optimal}) are used by the robot. $(v^*, \omega^*)$ ensures the robot's safety in terms of avoiding solid obstacles, and ProNav ensures walking stability and low power consumption.

%% file: 6_Results.tex
\section{Results and Analysis}
This section outlines ProNav's implementation, our chosen evaluation parameters, the varied test environments, and comparisons with other methods.

\subsection{Robot Setup and Dataset}
Our approach is implemented on a Spot robot, a 12-degree-of-freedom (DOF) quadruped from Boston Dynamics. The robot provides joint feedback from its 12 actuators and the battery current data during its operation. Our data collection was carried out at the University of Maryland College Park campus, on different terrains including concrete, asphalt, grass, rocks, sand, bushes, mulch, etc. The resulting dataset represents approximately 9 hours of operation, during which the joint feedback and battery current were continuously recorded. ProNav runs at 16 Hz on an Intel NUC edge computer equipped with an Intel i7 CPU, and an Nvidia RTX2060 GPU. 

\begin{figure*}[t]
    \centering
    \includegraphics[width=0.8\linewidth]{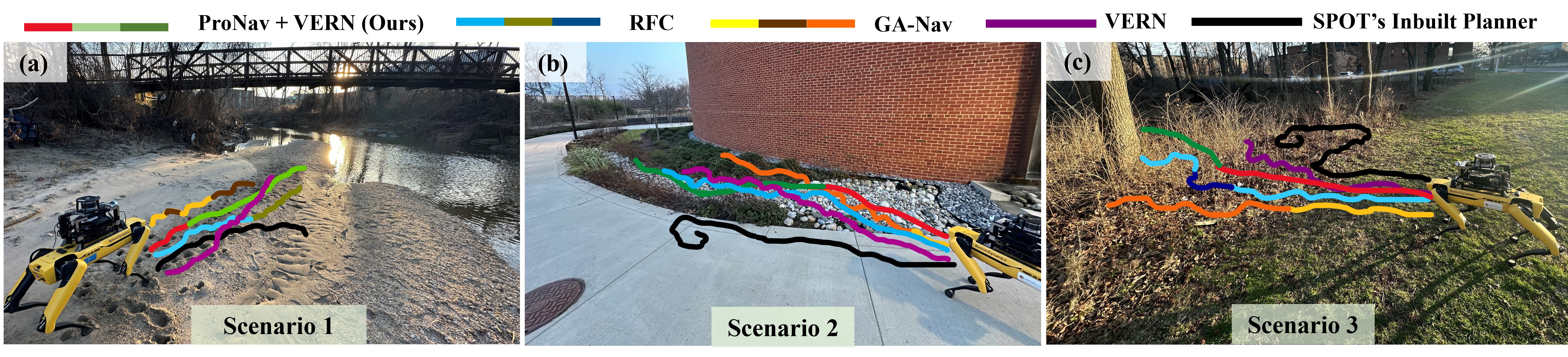}
    \caption{\small{An instance of Spot navigating in different outdoor terrains using: ProNav (trot: red, crawl: light green, amble: green), RFC (trot: light blue, crawl: army green, amble: dark blue),  
    GA-Nav (trot: yellow, crawl: brown, amble: orange), VERN (purple), Spot's in-built planner (black). 
    We observe that our method chooses the appropriate gait and velocities based on terrain traversability, leading to better success rates, lower power consumption, and vibration cost.}}
\label{fig:comparison}
\end{figure*}


\subsection{Comparison Methods}
We combine ProNav with VERN \cite{sathyamoorthy2023vern} and compare its performance with exteroceptive, and proprioceptive navigation techniques:

\begin{itemize}
    \item \textbf{Spot's in-built planner}: Uses $360^{\degree}$ RGB-D cameras to detect and avoid obstacles. It also automatically adapts the robot's leg raise heights based in the terrain.


    \item \textbf{VERN} \cite{sathyamoorthy2023vern}: Uses RGB images and 3D point clouds to differentiate pliable vegetation from solid obstacles and traverse vegetated environments.

    \item \textbf{GA-Nav} \cite{guan2022ga}: Uses RGB images 
    for semantic segmentation, computing traversability costs for various terrains. It is combined with a planner \cite{fox1997dynamic} to compute trajectories on low-cost terrains and avoid obstacles.

    \item \textbf{Random Forest Classifier (RFC)} \cite{kertesz2016rigidity}: Uses proprioceptive signals' input vector \( \mathbf{A}(t) \) to classify the terrain type. We used the dataset described in section V.A to train the classifier to identify between four different terrain types (granular, poor foothold, high resistance and stable). Also, RFC is combined with VERN to generate goal-directed velocities $(v^*, \omega^*)$.
    
\end{itemize}

\no We use GA-Nav \cite{guan2022ga} and RFC \cite{kertesz2016rigidity} to classify the terrains based on RGB images and proprioception, respectively. After that, we choose the appropriate gait according to the following:
Trot if the robot's trajectory is on stable terrains (concrete, asphalt), crawl for granular terrains (sand), and amble for terrains with poor footholds and high resistance to motion (forest, dense vegetation). Also, we also perform ablation studies by removing the joint positions and current components from our input to our PCA-based system.

\subsection{Evaluation Metrics}
\begin{itemize}
    \item \textbf{Success Rate}: The ratio of successful navigation trials where the robot was able to reach its goal without freezing or colliding with obstacles.
    
    \item \textbf{Mean power consumption}: 
    The amount of power (in Watts)consumed from the robot's battery averaged over all trials.
    
    \item \textbf{Mean Velocity}: The robot’s average velocity along its trajectory as it traverses various surfaces.

    \item \textbf{Time to Goal}:  The robot’s average time to reach its goal in the successful trials.
    
    \item \textbf{Vibration Cost}: 
    The cumulative sum of deviations in hip joint positions from a stable reference, measured at each time instant \( t \). Deviations for each hip joint \( j \) in the x- and y-axes are calculated as follows:
    {\small
    \[
    \delta_{j}(t) = 
    \begin{cases}
        \left| p^{j}(t) - \min(p^{j}_{\text{ref}}) \right| & \text{if } p^{j}(t) < \min(p^{j}_{\text{ref}}), \\
        \left| p^{j}(t) - \max(p^{j}_{\text{ref}}) \right| & \text{if } p^{j}(t) > \max(p^{j}_{\text{ref}}). \\
    \end{cases}
    \]
    }
    Here, \( p^{j}(t) \) is the position of joint \( j \) at time \( t \), and \( p^{j}_{\text{ref}} \) represents reference positions from a stable terrain (for this work concrete is considered a stable terrain). The total Vibration Cost at time \( t \) is then computed as \( \text{Vibration Cost}(t) = \sum_{j} \delta_{j}(t) \).
    

    \item \textbf{IMU Energy Density }: The mean of the aggregated squared acceleration values measured by the IMU sensors across the x, y, and z axes, calculated over the successful trials. The relevant equations implemented are adopted from \cite{try2023vibration}: 
    {\small
    \begin{equation}
    E_i = \sum_{n=1}^{N} a_i^2,
    \end{equation}
    \begin{equation}
    E_{\text{Total}} = E_{ax} + E_{ay} + E_{az},
    \end{equation}  
    }
    where \( a_i \) represents one of the three acceleration signals (x, y, and z axes), and \( N \) is the IMU readings along the trajectory.
\end{itemize}

\subsection{Test Scenarios}
\begin{itemize}
    \item \textbf{Scenario 1:} Granular terrains (small rocks and sand). 
    \item \textbf{Scenario 2:} Concrete, rocks, and vegetation. 
    \item \textbf{Scenario 3:} Sparse tall grass, fallen logs, and trees.
    \item \textbf{Scenario 4:} Dense vegetation and bushes.
\end{itemize}

\begin{table}
\caption{Our method's navigation performance, measured against other methods using four different metrics, shows ProNav excelling in success rate. Meanwhile, our method achieves the lowest power consumption and vibration cost.}
\centering
\scriptsize
\begin{tabularx}{\columnwidth}{|c|c|X|X|X|X|}
\hline
\textbf{Metrics} & \textbf{Method} & \textbf{Scen. 1} & \textbf{Scen. 2} & \textbf{Scen. 3} & 
\textbf{Scen. 4}\\
\hline
\multirow{6}{*}{\shortstack{Success\\Rate (\%)}} 
& In-built system  & 30 & 20 & 10 & - \\
& GA-Nav\cite{guan2022ga}  & 80 & 50 & 50 & 30\\
& VERN\cite{sathyamoorthy2023vern}  & 70 & 60 & 40 & 20 \\
& RFC\cite{kertesz2016rigidity}  & 80 & 70 & 70 & 50 \\
& w/o current+VERN  & 100 & 80 & 60 & \textbf{70} \\
& w/o position+VERN  & 90 & 70 & 80 & 60\\
& ProNav+VERN  & \textbf{100} & \textbf{90} & \textbf{90} & \textbf{70}\\
\hline
\multirow{6}{*}{\shortstack{Mean Power\\ Consumption\\(watts)}} 
& In-built system   & 503 & 462 & 542 & - \\
& GA-Nav\cite{guan2022ga}  & 384 & 373 & 374  & 442 \\
& VERN\cite{sathyamoorthy2023vern}  & 462 & 372 & 362 & 450\\
& RFC\cite{kertesz2016rigidity}  & 365 & 380 & 369 & 419\\
& w/o current+VERN  & \textbf{371} & 356 & 357 & 398 \\
& w/o position+VERN & 388 & 361 & 370 & 403\\
& ProNav+VERN & 379 & \textbf{349} & \textbf{353} & \textbf{375}\\
\hline
\multirow{6}{*}{\shortstack{Mean Velocity\\ (m/s)}}  
& In-built system   & 0.43 & 0.35 & 0.33 & - \\
& GA-Nav\cite{guan2022ga} & 0.29 & 0.43 & 0.33 & 0.35 \\
& VERN\cite{sathyamoorthy2023vern}  & 0.42 & 0.47 & 0.34  & 0.34 \\
& RFC\cite{kertesz2016rigidity}  & 0.30 & 0.39 & 0.35 & 0.33 \\
& w/o current+VERN  & 0.32 & 0.45 & 0.36 & 0.38 \\
& w/o position+VERN & 0.24 & 0.49 & 0.32  & 0.36 \\
& ProNav+VERN & 0.27 & 0.38 & 0.31  & 0.32 \\
\hline
\multirow{6}{*}{\shortstack{Time\\to\\Goal\\(seconds)}} 
& In-built system  & 20.13 & 25.11 & 25.17 & - \\
& GA-Nav\cite{guan2022ga} & 24.85 & 16.19 & 25.01 & 26.23 \\
& VERN\cite{sathyamoorthy2023vern} & 19.04 & 16.31 & 24.18 & 27.14\\
& RFC\cite{kertesz2016rigidity}  & 24.31 & 17.40 & 24.92 & 25.83\\
& w/o current+VERN & 26.82 & 18.45 &  23.42 & 22.78 \\
& w/o position+VERN  & 34.45 & 15.94 & 25.31  & 25.39 \\
& ProNav+VERN  & 25.68 & 21.02 & 27.73 & 25.97 \\
\hline
\multirow{6}{*}{\shortstack{Vibration\\Cost}} 
& In-built system  & 66.3 & 53.2 & 50.6  & - \\
& GA-Nav\cite{guan2022ga} & 19.6 & 14.9 & 29.5 & 30.2 \\
& VERN\cite{sathyamoorthy2023vern} & 23.4 & 25.5 & 27.8  & 33.1 \\
& RFC\cite{kertesz2016rigidity}  & 18.1 & 16.3 & 25.6 & 29.4 \\
& w/o current+VERN & 22.7 & 12.6 & 64.0 & 15.4 \\
& w/o position+VERN  & 39.4 & \textbf{3.1} & 23.7  & 12.8 \\
& ProNav+VERN  & \textbf{16.9} & 4.4 & \textbf{10.7}  & \textbf{7.1} \\
\hline

\multirow{6}{*}{\shortstack{IMU\\Energy Density}} 
& In-built system  & 55283 & 32367 & 46835 & - \\
& GA-Nav\cite{guan2022ga} & 41175 & 23919 & 19307 & 28366 \\
& VERN\cite{sathyamoorthy2023vern} & 51374 & 26950 & 17052  & 33948 \\
& RFC\cite{kertesz2016rigidity}  & 38957 & 25314 & 16329 & 30424 \\
& w/o current+VERN & 28075 & 16834 & 24499 & 24005 \\
& w/o position+VERN  & 25378 & 15223 & \textbf{13207} & 18750 \\
& ProNav+VERN  & \textbf{23503} & \textbf{12388} & 14274 & \textbf{17186} \\
\hline
\end{tabularx}
\label{tab:tab1}
\end{table}

\subsection{Analysis and Discussion}
In this section, we evaluate qualitatively and quantitatively the performance of our method and compare it with other methods. Figure \ref{fig:comparison} provides a visual representation of the trajectories in different terrains. Our method showcases its superiority in navigating through dense vegetation (Fig. \ref{fig:cover}), granular (Fig. \ref{fig:comparison}a), rocky (Fig. \ref{fig:comparison}b), and unstructured forested terrains (Fig. \ref{fig:comparison}c). Notably, our method adapts by choosing crawl on sand (scenario 1) and amble in other scenarios whenever poor footholds and resistance to motion dominate, and $\mathbf{p}_t \notin LVZ$. RFC exhibited effective terrain analysis using proprioception in the granular scenario. However, the classifier's performance declined in the other scenarios due to their unstructured nature. This complexity led to either frequent gait changes, or no gait change resulting in instability, or failure often caused by entanglements in vegetation. Spot's built-in system faces challenges in vegetation-rich scenarios (2, 3 and 4) due to its default trot gait leading to leg entanglement, as reflected in its lower success rates. It also exhibits unstable behavior when the ground contains branches, and rocks of various sizes, as it considers them as obstacles that should be circumvented. VERN also encounters failure instances due to leg entanglements, particularly in scenario 4 with denser vegetation. GA-Nav, similar to RFC shows efficiency in open and uncovered terrains like in scenario 1, but in vegetated scenarios (3 and 4), often struggles to accurately identify the correct terrain type, primarily due to motion blur caused by entanglements. Also, strong lighting (Scenario 3), and low lighting (Scenario 4) drastically affect VERN's and GA-Nav's performance. This leads to either frequent gait changes, or changing to an inappropriate gait (e.g. crawl in dense vegetation which causes further leg entanglement). 
Conversely, our method achieves the highest success rate in all scenarios with its appropriate gait adaption without requiring visual feedback. ProNav halts in extreme cases to prevent imminent crashes, particularly in dense vegetation scenarios (3 and 4).  Compared to the second best method, our method improves the success rate by 25\%, 28.57\%, 28.57\%, and 40\% in scenarios 1,2,3, and 4, respectively.

We compute the success rate improvement using the following equation:
\begin{equation}
\text{Improvement (\%)} = \left( \frac{SR_{\text{ours}} - SR_{\text{2nd}}}{SR_{\text{2nd}}} \right) \times 100
\end{equation}  
Where, \( SR_{\text{ours}} \) and \( SR_{\text{2nd}} \) represent the success rates of our method and the second-best method, respectively.

We note that our approach consistently yields the lowest power consumption in all evaluated terrains.  This efficiency is a result of its capability to assess stability and its superiority in gait selection. VERN, while comparable in certain scenarios, has increased power consumption in the fourth scenario due to its default trot gait leading to more entanglements and consequent motion resistance. Likewise, GA-Nav exhibits increased power consumption in scenario 4, primarily due to its multiple changes in gait selection.
Moreover, our method consistently records the lowest vibration levels (in terms of the vibration cost and the IMU energy density metrics). Conversely, the frequent changes in gait exhibited by RFC lead to increased vibration costs when traversing through dense vegetation. In scenario 4, RFC and GA-Nav show high vibration in scenario 4 due to entanglements and gait alternations. Also, Spot's in-built system 
experiences the highest vibration costs due to sinkage (scenario 1), slippage (scenario 2), and motion resistance (scenarios 3 and 4).
In the mean velocity metric, ProNav shows a reduced pace, particularly during the gait switch to crawl in scenario 1. 
ProNav's time to goal is comparable to other methods, except in scenario 1, where the exteroceptive-based methods used a faster, yet high vibration gait. 

\textbf{Ablation Study on Proprioceptive Signals:} Our ablation analysis focused on two proprioceptive signals of ProNav: current drawn from the battery and hip joints' positions. In our evaluations (Table \ref{tab:tab1}), omitting battery current resulted in notably delayed or incorrect traversability estimations, notably impacting power consumption and vibration costs, especially evident in scenario 4 in dense vegetation.  Removing hip joints' positions also hindered performance but to a lesser extent. Despite their relative performance, neither ablated configuration could exceed the performance of the fully integrated ProNav system. We did not remove the knee force for our ablation study, since a PCA cluster could not be formed without it, which hinders the comparison. 

Table \ref{tab:tab2} shows navigation comparisons when using a single stable gait (crawl or amble) as well as ProNav with its adaptive gait adjustment. We observe that crawl gait has the lowest power consumption and vibration levels compared to the amble gait. However, its application in dense vegetation presents challenges; the robot moves slowly, leading to its legs getting entangled with the vegetation. For instance, in scenario 4, we note elevated power consumption and vibration levels alongside a significantly low velocity.
In contrast, the amble gait consistently 
achieves superior velocities 
and reaches the goal quickly relative to crawl and ProNav. Also, it has high mean power consumption which reduces the risk of entanglement (as the robot exerts more torque), and consequently lower vibration cost. ProNav on the other hand provides the best trade-off between the average power consumption, vibration cost and mean velocity.

\begin{table}[htbp]
\caption{Navigation performance while using only stable gaits, crawl, and amble. ProNav adaptively selects the appropriate gait by assessing terrain traversability and stability.}
\centering
\small
\begin{tabularx}{\columnwidth}{|c|c|X|X|X|X|}
\hline
\textbf{Metrics} & 
\textbf{Method} & 
\textbf{Scen. 1} & 
\textbf{Scen. 2} & 
\textbf{Scen. 3} & 
\textbf{Scen. 4}\\
\hline
\multirow{3}{*}{\shortstack{Mean Power\\ Consumption\\(watts)}} 
& Crawl & 382 & 358 & 374 & 488\\
& Amble  & 443 & 370 &  421 & 427\\
& ProNav  & \textbf{379} & \textbf{349} & \textbf{353} & \textbf{375}\\
\hline\multirow{3}{*}{\shortstack{Mean Velocity\\ (m/s)}} 
& Crawl  & 0.29 & 0.24 & 0.21 & 0.17\\
& Amble  & 0.33 & 0.56 & 0.28 & 0.54\\
& ProNav & 0.27 & 0.38 & 0.31  & 0.32 \\
\hline
\multirow{3}{*}{\shortstack{Time to\\Goal\\(seconds)}} 
& Crawl  & 29.54 & 33.76 & 54.21 & 68.2 \\
& Amble  & 24.05 & 14.91 & 18.75 & 19.90 \\
& ProNav  & 15.2 & 2.7 & 8.1  & 4.8 \\
\hline
\multirow{3}{*}{\shortstack{Vibration cost}} 
& Crawl  & 18.3 & 34.4 & 46.2 & 37.8\\
& Amble  & 28.4 & 22.9 & \textbf{5.7} & \textbf{6.1}\\
& ProNav  & \textbf{16.9} & \textbf{4.4} & 10.7  & 7.1 \\
\hline
\multirow{3}{*}{\shortstack{IMU\\Energy Density}} 
& Crawl  & 18078 & 25194 & 22359 & 37405\\
& Amble  & 34901 & 18846 & \textbf{7892} & \textbf{15984} \\
& ProNav  & \textbf{23503} & \textbf{10388} & 14274 & 17186 \\
\hline
\end{tabularx}
\label{tab:tab2}
\end{table}



%% file: 7_Conclusion.tex
\section{Conclusion, Limitations \& Future work}
We present ProNav, a new method that uses proprioceptive data to evaluate terrain's traversability in real time for legged robots. Our method optimizes robotic gait selection for improved
stability and reduced energy consumption. Also, the inclusion
of an advanced crash prediction system ensures safer and more
efficient navigation. We also combined ProNav with an exteroceptive-based navigation method, which improved its performance. We validate our method in different outdoor environments 
and provide a detailed comparison with other navigational methods.

However, ProNav has some limitations. It can only assess the stability of the terrain the robot is currently on. This could lead to failures and crashes in extreme environments. To solve this, we are considering adding other sensor modalities (e.g. RGB, thermal, or hyperspectral images) that can provide meaningful lookahead for the robot. Our gait adaptation alternates between the existing gaits on our hardware platform as custom gaits cannot be executed on it. In the future, we would like to create and utilize custom gaits for stabilization on an open hardware platform.  We would also like to investigate techniques to improve crash prevention, adapting our approach to more diverse environments and situations where halting is insufficient to prevent a crash.